\pdfoutput=1

\documentclass[11pt]{article}

\usepackage{acl}

\usepackage{times}
\usepackage{latexsym}

\usepackage{graphicx}

\usepackage[T1]{fontenc}

\usepackage[utf8]{inputenc}

\usepackage{microtype}

\title{A Robust Bias Mitigation Procedure Based on the Stereotype Content Model}

\author{Eddie L. Ungless \and Amy Rafferty \and Hrichika Nag \and Björn Ross \\
        School of Informatics \\ University of Edinburgh \\ 10 Crichton Street, Edinburgh EH8 9AB, United Kingdom \\
        \texttt{e.l.ungless@sms.ed.ac.uk a.rafferty@live.com} \\ \texttt{naghrichika@gmail.com b.ross@ed.ac.uk}}

\begin{document}
\maketitle
\begin{abstract}
The Stereotype Content model (SCM) states that we tend to perceive minority groups as cold, incompetent or both. In this paper we adapt existing work to demonstrate that the Stereotype Content model holds for contextualised word embeddings, then use these results to evaluate a fine-tuning process designed to drive a language model away from stereotyped portrayals of minority groups. We find the SCM terms are better able to capture bias than demographic agnostic terms related to pleasantness. Further, we were able to reduce the presence of stereotypes in the model through a simple fine-tuning procedure that required minimal human and computer resources, without harming downstream performance. We present this work as a prototype of a debiasing procedure that aims to remove the need for \textit{a priori} knowledge of the specifics of bias in the model.
\end{abstract}

\section{Introduction}

\label{sec:intro}
It is well established that large language models (LLMs) such as BERT \cite{Devlin_Chang_Lee_Toutanova_2019}, GPT2 \cite{Radford_Wu_Child_Luan_Amodei_Sutskever} and related contextualised word embeddings such as ELMo \cite{peters-etal-2018-deep} are biased against different demographic groups \cite{Guo_Caliskan_2021,Webster_Wang_Tenney_Beutel_Pitler_Pavlick_Chen_Chi_Petrov_2021,Kaneko_Bollegala_2021}, in that they often reflect stereotypes in their output. For example, given the prompt ``naturally, the nurse is a", these systems will typically output ``woman'' \cite{Schick_Udupa_2021}. Given the common practice of adapting pre-trained language models for a range of tasks through fine-tuning, upstream bias mitigation may prove to be the most efficient solution \cite{Jin_Barbieri_Kennedy_Davani_Neves_Ren_2021} (though cf. \citet{Steed_Panda_Kobren_Wick_2022}. In this paper, we demonstrate the success of modifying an existing debiasing algorithm to be grounded in a psychological theory of stereotypes - the SCM \cite{Cuddy_Fiske_Glick_2008}, to efficiently reduce biases in LLMs across a range of identities. Our proposed debiasing pipeline has the benefit of minimising the time spent researching identity terms and associated stereotypes. Being a fine-tuning procedure, this also reduces the amount of computational resources needed compared to training an unbiased model from scratch. This renders our approach efficient and widely applicable. We demonstrate using BERT, but this same procedure could easily be adapted to other LLMs. 

We adapt the fine-tuning procedure from \citet{Kaneko_Bollegala_2021}. They reduce gender bias in a range of LLMs by fine-tuning using a data set of sentences containing (binary) gendered terms (like ``he, man'' or ``she, lady'') (which they call attributes) or stereotypes associated with different genders (``assertive, secretary'') (which they call targets). The training objective is to remove associations with gender in the contextualised embeddings of the targets whilst maintaining these associations for the gendered attributes. 

Crucially, rather than relying on stereotypes specific to a particular demographic such as men and women (as in \citet{Kaneko_Bollegala_2021}) we plan to use the SCM to inform our production of fine-tuning data, inspired by work by \citet{Fraser_Nejadgholi_Kiritchenko_2021}. The SCM states that our stereotyped perception of different demographics can be conceptualised as lying in a vector space with axes of warmth/coldness and competence/incompetence \cite{Cuddy_Fiske_Glick_2008}. We tend to consider our own identity group to be warm and competent, and stereotype disfavoured groups such as people experiencing homelessness as cold and/or incompetent \cite{Cuddy_Fiske_Glick_2008}. 

In the terminology of \citet{Kaneko_Bollegala_2021}, our attributes are terms relating to warmth and competence taken from \citet{Nicolas_Bai_Fiske_2021} (as in \citet{Fraser_Nejadgholi_Kiritchenko_2021}, a paper on stereotypes in static embeddings), our targets are demographic identity terms. Because the SCM is designed to encompass many different minority groups, this avoids the need to generate lists of stereotypes unique to each minority group, reducing work load and making the tool easy to adapt to different targets. Therefore, the procedure should be effective for all identity terms we use. We demonstrate this technique for Black/white ethnicity and also the intersectional power dynamic between white men and Mexican American women, but this could easily be expanded to other aspects of identity such as disability and sexuality. Further, whilst we focus on English language and American identities, there is evidence that the SCM may hold relatively well cross-culturally \cite{Cuddy_Fiske_Kwan_2009}, so this approach may be transferable to other LLMs. 

We adapt the Contextualised Embedding Association Test (CEAT) \cite{Guo_Caliskan_2021} using the vocabulary from \citet{Nicolas_Bai_Fiske_2021} in order to measure stereotypes in contextualised word embeddings.
The CEAT provides a robust measure of bias in contextualised word embeddings for target words, and is suited for use with the SCM terms.

In addition to using the CEAT to test for bias, we also measure the performance of the model on the language modeling benchmark GLUE \cite{wang2019glue}, to ensure the fine-tuning procedure does not adversely impact the quality of the model, an issue \citet{Meade_Poole-Dayan_Reddy_2021} identify as affecting several debiasing techniques. 

The main contributions of this paper are to demonstrate:
\begin{itemize}
\setlength\itemsep{-0.2em}
    \item that the SCM can be used to detect bias in contextualised word embeddings
    \item a debiasing procedure that is demographic agnostic and resource efficient\footnote{Code available at \url{https://github.com/MxEddie/Demagnosticdebias}}  
\end{itemize}

\section{Related work}\label{sec:relwork}
Several contributions have been made towards measuring and mitigating bias in NLU models with minimal \textit{a priori} knowledge. Fraser and colleagues (\citeyear{Fraser_Nejadgholi_Kiritchenko_2021}) demonstrated the validity of the SCM for static word embeddings, in that the embeddings of words associated with traditionally oppressed minority groups such as Mexican Americans or Africans tend to lie in the cold, incompetent space, as determined by cosine similarity. Note that, unlike \citet{Fraser_Nejadgholi_Kiritchenko_2021}, we focus on the embeddings of the identity terms themselves, not of words associated with those identities, as we explicitly want to identify whether there is bias in the embeddings. \citet{Fraser_Nejadgholi_Kiritchenko_2021} looked to establish if the embeddings of associated terms followed the SCM's predictions, not whether the word embeddings were biased in a way as to reflect these stereotypes. 

\citet{Utama_Moosavi_Gurevych_2020} propose a strategy for debiasing ``unknown biases''. They train a shallow model which picks up superficial patterns in data that are likely to indicate bias. This is then used to train the main model, which works by downweighting the potentially biased examples, paired with an annealing mechanism which prevents the loss of useful training signals caused by this approach. The models obtained from this self-debiasing framework were shown to perform just as well as models debiased using prior knowledge. In our work we do not train our model from scratch and only focus on social bias, whereas \citet{Utama_Moosavi_Gurevych_2020} do not target specific bias types. We chose to prioritise socially relevant biases with the hopes of minimising harm done to minority communities. Further, our method requires far less compute. 

\citet{Webster_Wang_Tenney_Beutel_Pitler_Pavlick_Chen_Chi_Petrov_2021} take gendered correlations in pretrained language representations as a case study for measuring and mitigating bias. They build an evaluation framework for detecting and quantifying gendered correlations in models. They find that both dropout regularization and counterfactual data augmentation minimize gendered correlations while maintaining strong model accuracy. Their techniques are applicable when training a model from scratch, whilst ours is a fine-tuning procedure, meaning it requires fewer computational resources. 

\citet{Schick_Udupa_2021} explore whether language models can self-diagnose undesirable outputs for self-debiasing purposes. Their approach encourages the model to output biased text, and uses the resulting distribution to tune the model's original output. We argue that our model is more demographic agnostic, as their approach depends heavily on biases captured by Perspective API. Their approach may miss less salient forms of bias as it relies on the model having some representation of the bias category beforehand. Using the SCM, we can work ``backwards'' from the fact that these communities are harmed to then assume they will be represented as cold and/or incompetent, making our approach more universally applicable. 

\citet{cao-etal-2022-theory} focuses on identifying stereotyped group-trait associations in language models, by introducing a sensitivity test for measuring stereotypical associations. They compare US-based human judgements to language model stereotypes, and discover moderate correlations. They also extend their framework to measure language model stereotyping of intersectional identities, finding problems with identifying emergent intersectional stereotypes. Our work is unique from this in that we have additionally performed debiasing informed by the SCM.

Overall, our methodology and approach differs from most other contributions in this field as it focuses on targeting social bias specifically, and we propose a fine-tuning debiasing approach which requires little in the way of human or computer resources and is not limited to a small number of demographics.

\section{Data sets and tasks} 
\subsection{Data for Debiasing Procedure}
\subsubsection{Identity terms (targets)}\label{identity_terms}
We established two sets of identity terms (targets) for use with the context debiasing algorithm. The first set relates to racial bias (bias against people of colour based on their (perceived) race). BERT has been shown to demonstrate racial bias in both intrinsic \cite{Guo_Caliskan_2021} and extrinsic measures \cite{Nadeem_Bethke_Reddy_2020, Sheng_Chang_Natarajan_Peng_2019}. To reduce bias against Black people compared to white, we created a list of 20 African American (AA) and 20 European American (EA), 10 male and 10 female names for each, to use in the debiasing procedure. We used names from \citet{Guo_Caliskan_2021} (excluding any included in the CEAT tests we deploy, see Section \ref{ceat}) and supplemented these lists with common names from a database of US first names \cite{Tzioumis_2018}. Excluding names from the CEAT tests was crucial to ensure a reduction in bias was due to a restructuring of the embedding space and an overall change in how Black individuals were represented, and not due to bias reduction for the specific names we ran the debiasing procedure with. 

The second set relates to intersectional bias against Mexican American (MA) women, that is bias against women based on both patriarchal beliefs about their gender and prejudice against their ethnicity. This intersectional bias is evident in the contextualised embeddings BERT produces \cite{Guo_Caliskan_2021}. To reduce bias against MA women compared to white men, we additionally took 10 common Hispanic female names (and manually confirmed that each was used by the Mexican American community through a Google search) from \citet{Tzioumis_2018}. 

The validity of using names to represent demographic groups has been questioned \cite{Blodgett_Lopez_Olteanu_Sim_Wallach_2021}. However, we assume that reducing bias present in the representations of these names will go some way to reducing racial bias in the model. 


\subsubsection{Stereotype Content terms (attributes)}\label{attributes}
As with \citet{Fraser_Nejadgholi_Kiritchenko_2021}, we use the Stereotype Content terms from \citet{Nicolas_Bai_Fiske_2021}, whereby the high morality, high sociability terms are taken to indicate warmth; low morality, low sociability to indicate coldness; high ability, high agency to indicate competence; and low ability, low agency to indicate incompetence. We selected the top 32 most frequent terms from each list (as measured using the Brown Corpus and the NLTK toolkit), to increase the likelihood we would find a large number of example sentences for each. During finetuning, we wish for these terms to maintain their projection in the warmth/coldness or competence/incompetence space, respectively, whilst removing projection in these directions for the target terms (see Section \ref{method} and Figure \ref{fig:projection}).

Whilst the exact ``position'' of demographic groups in this conceptual space would vary depending on who is describing them, in this work we always assume the minority group will be represented in the original model as cold and incompetent, in other words the most disfavoured and most likely to experience harm \cite{Cuddy_Fiske_Glick_2008}. This minimises workload (no need to establish likely predictions for every demographic considered, beyond identifying the more marginalised group) and centers our approach around improving results for the most negatively represented identity terms. Note, there is no harm in running our debiasing procedure on identities that are already equally associated with one concept i.e. warmth, whilst also reducing stereotyped associations with the other concept i.e. competence. 

\subsubsection{Fine-tuning data}
Having established the list of attribute and target terms, we follow an adapted version of \citet{Kaneko_Bollegala_2021}'s procedure for generating fine-tuning development data. During early analyses, we found the AA names occurred very infrequently in their provided news commentary data set, likely a reflection of the lack of AA representation in mainstream news \cite{Diuguid_Rivers_2000}. We therefore opted to use data from Reddit, from 2018\footnote{\url{https://files.pushshift.io/reddit/comments/}}, (a separate data set to that used for the CEAT, see below), as this contained many example sentences across all names. We sampled from this data set sentences which contained either one of the attribute or one of the target terms, and no more, of 128 tokens or less. We extracted at least 24,000 sentences for each attribute and target dimension. This was stored as a dictionary that was passed to the debiasing script. We took a random sub-sample of 1000 of each to use as development data. 

\subsection{CEAT}\label{ceat}
The CEAT \cite{Guo_Caliskan_2021} is designed to test for associations between the contextualised embeddings of targets and polar attributes (such as binary gender). The authors sampled sentences from Reddit where a stimuli (target or attribute term) occurred, and generated contextualised embeddings for the sentences. These contextualised embeddings were then used to calculate the effect sizes, based on a cosine similarity measure between the embeddings of the target and attribute tokens. They then measure the distribution of effect sizes for the terms in different contexts (to ensure that the choice of context does not unduly influence the final effect size metric). The authors then apply a random-effects model to calculate a combined effect size (CES) and significance, given the distribution of effect sizes. We adopt the same sample data and testing procedure. 

We use the lists of identity terms for racial and intersectional bias given in \citet{Guo_Caliskan_2021}, namely related to AA versus EA identities and MA women versus EA men, along with the SCM attribute terms, to establish the presence of stereotypes in the contextualised word embeddings using the CEAT. 

In addition to using the SCM terms, we will also use the pleasant/unpleasant terms from \citet{Guo_Caliskan_2021}'s paper - this provides a comparison point for use of the SCM versus another set of non-demographic-specific terms. 

We also measure how strongly the demographic specific stereotype terms for MA women and EA men are associated with the demographic groups, to see if demographic specific stereotype associations are reduced following demographic agnostic debiasing. Note that we removed the word "intelligent" from the EA men attributes list as this also occurs in the competence attributes list and we wanted to be totally confident that any observed reduction in bias was due to restructuring of the entire embedding space and not due to bias being removed from an overlapping word. The CEAT does not have equivalent demographic specific terms for the AA/EA groups, though for completeness we compare how strongly the MA female/EA male specific terms are associated with the AA/EA groups. 

Again, we adopt the approach of always assuming the more marginalised group will be represented in the model as more cold and incompetent compared to the majority group. This is an oversimplification. For example, \citet{Cuddy_Fiske_Glick_2008} indicate that in a Western context neither men nor women are strongly associated with coldness. However, we adopt this simplifying assumption to maintain testing consistency and thus require less human intervention, as per our goals. 

We apply the CEAT before and after debiasing, to measure the success of the fine-tuning approach using the SCM terms. 

\subsection{Language Modelling Benchmark}\label{glue_description}

\citet{Meade_Poole-Dayan_Reddy_2021} note that apparent reductions in bias can reflect a worsening of language modelling performance. To ensure our debiasing procedure does not come at the expense of model performance, we evaluate our model on the GLUE benchmark \cite{wang2019glue}.

The GLUE benchmark consists of 10 primary tasks and one diagnostic test, which evaluate the performance of a model in different contexts. We chose to evaluate our models using only five of these tasks -- MRPC, SST-2, STSB, RTE and WNLI -- following \citet{Kaneko_Bollegala_2021}. These five tasks have small datasets, meaning we can minimise the effect of task-specific fine-tuning when running predictions \cite{Kaneko_Bollegala_2021}. 

We run the tests using the public GLUE code from huggingface\footnote{\url{https://github.com/huggingface/transformers/tree/main/examples/pytorch/text-classification/}}. We will perform these tests before and after debiasing, and compare the results. We report results based on the provided evaluation data.

\section{Methodology}\label{method}
We use the `bert-base-cased' model from the Hugging Face library\footnote{\url{https://huggingface.co/bert-base-cased}}), henceforth \textsc{bert}, although this same procedure should be applicable to any LLM with minimal modification. 

\begin{figure}[t]
\begin{center}
{\centerline{\includegraphics[width=\columnwidth]{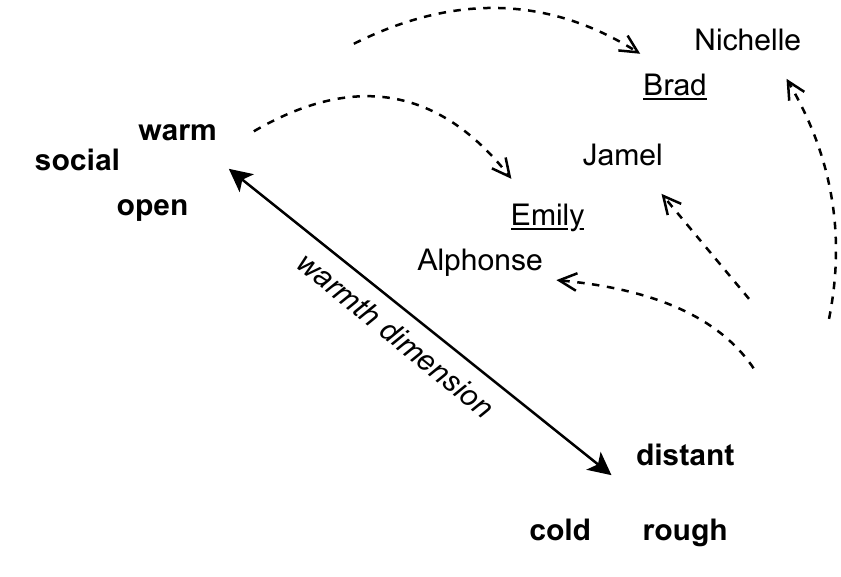}}}
\caption{Diagram of intended orthogonal projection of target terms away from the warmth dimension, determined by attribute terms in \textbf{bold}. EA names \underline{underlined}}. 
\label{fig:projection}
\end{center}
\end{figure}

We fine-tune the model following an adapted version of the procedure in \citet{Kaneko_Bollegala_2021}. Namely, through a training objective that looks to minimise unwanted projection into the attribute category dimensions for the target words through an orthogonal projection, whilst also staying close to the contextualised embeddings of the pre-trained model to preserve semantics. We visualise this orthogonal projection in Figure \ref{fig:projection}.
Adjusting the embeddings of the target terms to lie orthogonal to the warmth dimension (equidistant from the attribute terms) should ensure less negatively biased representations for minority groups (in the visualisation, AA names). 

Crucially, we modified the original algorithm in \citet{Kaneko_Bollegala_2021} as we wish to remove unwanted projections into two dimensions, not just one: warmth/coldness and competence/incompetence. The first component of the loss function for layer $i$ of our model is:

$$ L_{i} = \sum\limits_{d \in D}\sum\limits_{t \in V_{t}} \sum\limits_{x \in \Omega(t)}\sum\limits_{a \in V_{a}}(v_{i}(a)^{\top}E_{i}(t;x;\theta_{e}))^{2} $$


where $E_{i}(t;x;\theta_{e})$ represents the embedding of target word $t$ in sentence $x$ for model $E_{i}$, $v_{i}(a)$ is the average embedding for the attribute term across training sentences, and we calculate the inner product across all attributes $a \in V_{a}$, for all sentences containing the target $x \in \Omega(t)$, for all target words $V_{t}$, for all target dimensions, $D_{d}$.

The second component of the loss function is:

$$ L_{reg} = \sum\limits_{x \in A} \sum\limits_{w \in x}\sum\limits^{N}_{i=1}||E_{i}(w;x;\theta_{e})-E_{i}(w;x;\theta_{pre})||^{2} $$

where $E_{i}(w;x;\theta_{pre})$ is the contextualised embedding of a word, $w$, in a sentence, for the model before fine-tuning, and we calculate the squared $\ell_{2}$ between this and the embedding after fine-tuning, for all layers, for all sentences and targets. 

The final loss function is a weighted sum:
$$L = \alpha L_{i} + \beta L_{reg}$$
where $\alpha$ and $\beta$ sum to 1. 

\citet{Kaneko_Bollegala_2021} find debiasing all layers to be the most effective, so we do likewise.

\section{Results}
\label{sec:expts}
\subsection{Baseline Performance}
\subsubsection{CEAT}\label{ceat-baseline}
Results for the CEAT for \textsc{bert} are given in Table \ref{tab:ceat_table}. We found there was a medium combined effect size (CES, between 0.5 and 0.8, as per the original paper's classification \cite{Guo_Caliskan_2021}) in the strength of association between EA names \& warmth and AA names \& coldness. We also found a medium strength association between EA names \& competence and AA names \& incompetence. 
As with the original paper, we found a small association between EA names \& pleasantness and AA names \& unpleasantness, suggesting this approach may be less able to detect the true scale of bias. 

We also found a medium effect size association between AA names and the negative, MA women specific intersectional bias terms, and between the EA names and the EA male specific intersectional bias terms. This may be because the EA male stereotypes are relevant to all EA people. 

For the intersectional power dynamic, we found a small association between EA male names \& warmth and MA female names \& coldness. We found a medium association between EA male names \& competence and MA female names \& incompetence. We found a very small association between EA male names \& pleasantness and MA female names \& unpleasantness, suggesting these generic terms are less effective for detecting the true levels of bias in the model. 

Finally, we found a medium effect size association between MA female names and the MA female specific bias terms, and between the EA male names and EA male specific bias terms - surprisingly, this association was weaker than than for the black/White demographic group, despite the fact that these stereotypes were chosen to be highly pertinent to the intersectional group. 

\begin{table*}[tb]
\begin{center}
\begin{tabular}{lrrrr}
\hline
 & \multicolumn{2}{c}{\textsc{bert}} & \multicolumn{2}{c}{\textsc{debias}} \\
Test & CES & Sig. & CES & Sig.  \\
\hline
EA,AA,Warm    & 0.77 & * & \textbf{-0.12} & - \\
EA,AA,Comp. & 0.67 & * & \textbf{-0.18} & - \\
EA,AA,Pleas.    & 0.47 & * & \textbf{0.16} & * \\
EA,AA,Inter.$^\dagger$ & 0.71 & * & \textbf{0.15} & * \\
EAM,MAF,Warm    & 0.43 & * & \textbf{-0.03} &  -    \\
EAM,MAF,Comp.     & 0.51 & *  & \textbf{-0.04} & - \\
EAM,MAF,Pleas.      & 0.17 & *  & \textbf{0.13} & *  \\
EAM,MAF,Inter. & 0.50 & * & \textbf{0.08} & *  \\
\hline
\end{tabular}
\caption{Strength of combined effect size (CES) between attributes and targets for BERT before (\textsc{baseline}) and after (\textsc{debiased}) debiasing. Sig. = significance. * = significant to $p < 0.05$. AA = African American names. EA = European American names. MAF = Mexican American female names. EAM = European American male names. Warm = warm/cold terms. Comp. = competent/incompetent terms. Pleas. = pleasant/unpleasant terms. Inter = Intersectional stereotypes.$^\dagger$ \textbf{Bold} indicates that the debiasing procedure has reduced the absolute effect size to very small.
$^\dagger$The intersectional stereotypes were intended as relevant to the EAM and MAF pair.}
\label{tab:ceat_table}
\end{center}
\end{table*}

\subsubsection{GLUE}
Table \ref{tab:glue} shows the GLUE benchmark scores for \textsc{bert} and \textsc{debias}, on the five chosen tasks.

The baseline \textsc{bert} model performs very well on SST-2, MRPC and STS-B, with metric scores of around 90\%. The lower scores come from the RTE and WNLI tasks. RTE assesses the model’s ability to determine whether sentence A entails sentence B. WNLI assesses the model’s ability to determine whether an inserted noun is correct. These specific grammatical situations seem to be the weaknesses of the model. The low score for WNLI is surprising and may indicate suboptimal hyperparameter choices during training. The training loss is comparable to that of a similar model on huggingface\footnote{\url{https://huggingface.co/gchhablani/bert-base-cased-finetuned-wnli}}. 

\begin{table}[t]
    \centering
    \begin{tabular}{lrr}
         \hline
         Benchmark & Baseline Score & Debiased Score \\
         \hline
         SST-2 & \textbf{92.7} & 92.5 \\
         MRPC & \textbf{89.5/85.0} & 87.9/82.8 \\
         STS-B & \textbf{88.9/88.6} & 88.7/88.5 \\
         RTE & 66.1 & \textbf{67.5} \\
         WNLI & 32.4 & \textbf{42.3} \\
         \hline
    \end{tabular}
    \caption{GLUE Benchmark scores for both our baseline \textsc{bert}, and our final \textsc{debias} models. Values correspond to the metrics described in Section \ref{glue_description}. \textbf{Bold} indicates the best performance.}
    \label{tab:glue}
\end{table}

\subsection{Debiasing Procedure}
We adopt the values for $\alpha$ and $\beta$ given in the original paper, namely 0.2 and 0.8 respectively, having trialed  $\alpha$ 0.1 above and below and found 0.2 to be the best performing. Bar batch size and learning rate, all other hyperparameters were set to their default values for \textsc{bert}. We trialed a number of starting learning rates and found the best to be 5e-5 (this is the same learning rate used in the original paper). Batch size was set to 32, as in the original paper. We train for 3 epochs (this is given in the code for the context debias paper but not specified). 

We fine-tuned the model using the methodology detailed in Section \ref{method}.

\subsection{Post-debiasing Performance}
\subsubsection{CEAT}
The results of our post-debiasing CEAT tests indicate this debiasing procedure to be largely successful. We were able to reduce bias in \textsc{debias} and in all instances render the strength of stereotyped association to be very small.

For \textsc{debias}, there is no longer an association between EA names \& warmth and AA names \& coldness, nor between EA names \& competence and AA names \& incompetence. 
Although our debiasing procedure involved only the SCM terms, it also had an impact on the other associations. The strength of association between EA names \& pleasantness and AA names \& unpleasantness has reduced to be very small. Intersectional bias was also reduced as to be very small. Though these very small effects are statistically significant, their practical impact will be negligible.  

Similarly, we found that for \textsc{debias}, there is no longer an association between EA male names \& warmth and MA female names \& coldness, nor between EA male names \& competence and MA female names \& incompetence. The association with pleasantness was also reduced, although this effect size was very small to begin with. Intersectional bias was also reduced as to be very small. 
\subsubsection{GLUE}
Table \ref{tab:glue} shows the differences between GLUE benchmark scores for our model before and after debiasing. For most tests, the GLUE benchmark scores have very minor differences.

Our debiased model outperforms the baseline model on both the RTE and WNLI tasks, with the largest difference coming from WNLI. We suspect that the improvement regarding RTE is because the RTE dataset is constructed based on news and Wikipedia text \cite{wang2019glue}, which are domains likely to contain significant bias. For WNLI, the task of resolving ambiguities requires real world knowledge, which is also highly influenced by bias. Removing bias from these datasets allows the model to focus on classifying entailment (RTE) or resolving ambiguities (WNLI) in a more reliable manner, without being ``distracted'' by stereotyped associations between particular groups and actions that are irrelevant to the task.

In general, these results show that debiasing the model did not hurt its performance, as would have been implied by \citet{Meade_Poole-Dayan_Reddy_2021}. On our five chosen GLUE tasks, any performance decreases were very minor, while the performance increases on RTE and WNLI were rather significant. Though not directly comparable to \citet{Kaneko_Bollegala_2021}, as their paper considers `bert-base-uncased', our results are inline with their findings showing debiasing along two ``axes'' does not unduly harm language modeling performance compared to debiasing along one axis. 

\section{Discussion}
We found that our approach to bias measurement, informed by the SCM, proved to be an effective method for detecting bias in an LLM. We found that compared to using another list of generic, non-demographic specific attribute terms related to pleasantness, our approach seemed to give a more accurate measure of the level of bias in the model - our terms allow us to capture a stronger association between a minority group and negative stereotypes. It is possible that our approach exaggerates the level of bias in the model and in fact is less accurate. However, the effect sizes from our approach are closer to the effect size for association with demographic specific terms for the intersectional pair, suggesting it paints an accurate picture of negative bias in the model. Further, given how often BERT has been found to produce offensive content, it seems more likely that use of pleasantness terms is underestimating the level of bias in the model, rather than our approach overestimating it. The pleasantness terms were only slightly associated with EA male names compared to MA female names, yet BERT has been shown to consistently produce more favourable content about such individuals \cite{Sheng_Chang_Natarajan_Peng_2019}.

Our finding that the intersectional bias terms were actually more strongly associated with the Black/white demographic groups highlights how the selection of demographic specific stereotypes for use in measuring bias and debiasing models can be challenging. That these stereotypes are actually more strongly associated with AA/EA names could suggest that the stereotyping captured by the model does not reflect the attitudes of the group of undergraduates responsible for generating these stereotypes \cite{Ghavami_Peplau_2013}. It could also be that the model has not been exposed to sufficient (stereotyped) data to capture the category of MA females and the associated stereotypes. 

The results might suggest that these demographic specific terms are actually rather ``demographic agnostic'', hence they are able to capture bias against AA people. However, intuitively, ``sexy'' and ``feisty'' (two MA female specific stereotypes) are not associated with people experiencing homelessness (and studies on public attitudes towards homelessness to our knowledge confirm this intuition), but the Stereotype Content Model is able to predict the contempt they experience due to being perceived as cold and incompetent \cite{Cuddy_Fiske_Glick_2008}, which is likely reflected in language use and thus in an LLM.  

After debiasing using the SCM informed approach, we were able to reduce bias in all instances. Not only did we reduce the association between competence, warmth and ethnicity, but we also reduced the association with pleasantness. Intuitively, this is likely a reflection of the semantic association between warmth and pleasantness - reducing projection in the warmth dimension may have impacted projection in the pleasantness dimension.

Crucially, we were able to reduce the association between the intersectional groups and their specific stereotypes, using a demographic agnostic approach that did not require prior knowledge of group specific stereotypes. Although we only ran the debiasing procedure for warmth and competence dimensions, there was a positive ``knock on'' effect, supporting our belief that debiasing at the more abstract level will reduce more specific bias associations as well, as these can be thought of as subcategories of these more generic stereotype concepts. We were able to successfully debias the model without impeding performance on benchmark NLI tasks, suggesting language modelling abilities have not been negatively impacted, and in two instances performance was actually improved, possibly due to the reduction in bias. 

\section{Conclusions}
\label{sec:concl}
\subsection{Future Work and Limitations}
In future work we hope to make use of language models to generate the target identity terms, akin to \citet{schick_dino}'s use of LLMs to generate training data, using prompts such as ``I am proud to identify as''. This will further reduce the amount of human resource and \textit{a priori} knowledge needed, making the approach more efficient and widely applicable. We may also try to introduce additional dimensions related to ``universal'' patterns of discrimination such as the use of dehumanising language \cite{Cameron_Harris_Payne_2016} and animal comparisons \cite{Haslam_Loughnan_Sun_2011}. 

Though we are hopeful that our proposed debiasing pipeline will show promising results, we acknowledge there are several inherent limitations we would look to address in future work.

First, the SCM has received significant support as a model for our perceptions of different groups, and its simplicity makes it ideal for use in our ``demographic agonostic'' approach. However, it has been shown that the model may fail to adequately capture stereotypes surrounding immigrant groups \cite{Greenwood_Blankenship_Stewart_Deaux_2021}. This might be addressed in future work by adopting additional attribute dimensions (i.e. diligence) to encompass a wider range of potential stereotypes. This will allow us to better measure and mitigate bias against groups which is not best captured by the warmth and competence stereotypes. 

A second limitation is our use of Reddit data for both debiasing and testing for bias - it is not clear how robust the reduction in bias would be if tested using out-of-domain data.

A further limitation is that during the process of identifying suitable names from \citet{Tzioumis_2018} for our debiasing procedure, we found that some of the names used in CEAT tests to measure bias against Black Americans were not predominantly used by Black individuals (for example ``Leroy''), an indication that relying on names to establish bias against a demographic group may be fallible.

Our use of the GLUE metric to evaluate language modelling performance is potentially problematic as this static benchmark is outdated and saturated for some tasks. Though using the same metric as \citet{Kaneko_Bollegala_2021} gave us confidence that debiasing along two axes did not unduly harm performance, we could better evaluate our model using modern dynamic benchmarks. 


Finally, intrinsic measure of bias do not always correlate well with application bias \cite{GoldfarbTarrant2021,Cao_Pruksachatkun_Chang_Gupta_Kumar_Dhamala_Galstyan_2022}, suggesting the CEAT may not accurately capture the extent of bias the model might be responsible for in downstream applications. In future work, we could evaluate the success of our debiasing approach using gender targets and an extrinsic measures such as \citet{winobias}, a gender bias in coreference resolution benchmark that could assess our model after finetuning for this task. We could also try to adapt the principles of this process to work in downstream tasks, for example amending the finetuning data to contain balanced stereotyped instances.

\subsection{Conclusion}
Our debiasing procedure has reduced stereotyped associations between minority groups and negative characteristics without the need for idiosyncratic target terms for each group, making it demographic agnostic and human resource efficient, in line with our goals. The debiasing procedure is able to effectively ``neutralise'' the presence of target dimensions in the attribute embeddings, as well as decrease the association between more demographic specific stereotype attributes and the target demographics. The debiasing procedure did not come at the cost of performance, and even improved performance on RTE and WNLI.

Further, the finetuning procedure ran in a matter of hours on a single GPU, making it computationally efficient as well. This aligns with our goals, to establish a robust bias mitigation procedure that is efficient and widely applicable. 

Our work can be thought of as a prototype for a promising debiasing procedure grounded in the SCM. In future, we hope to encompass automatic target term generation. We also plan to expand this work to more minority identities, and more importantly test the resulting model using a range of extrinsic bias measures and language modeling benchmarks, to evaluate the potential for a positive real world impact. The hope is that those using LLMs may apply our simple and efficient debiasing procedure before fine-tuning for their own purposes, helping to reduce the impact of stereotypes across the field. 

\section{Acknowledgements}
We would like to thank our anonymous reviewers for their feedback. This work is in part supported by the UKRI Centre for Doctoral Training in Natural Language Processing, funded by the UKRI (grant EP/S022481/1) and the University of Edinburgh, School of Informatics.

\bibliography{custom}
\bibliographystyle{acl_natbib}

\appendix



\end{document}